\documentclass[letterpaper, 10 pt, conference]{ieeeconf}  % Comment this line out
\IEEEoverridecommandlockouts                              % This command is only
\usepackage{multirow}
\usepackage{pifont}
\usepackage[utf8]{inputenc}
\usepackage[T1]{fontenc}
\usepackage{amsmath}
\usepackage{booktabs}
\usepackage[table]{xcolor}
\definecolor{rblue}{rgb}{0,0.5,1}
\definecolor{awesome}{rgb}{1.0, 0.13, 0.32}
\definecolor{hollywoodcerise}{rgb}{0.96, 0.0, 0.63}
\definecolor{lasallegreen}{rgb}{0.03, 0.47, 0.19}
\definecolor{hanpurple}{rgb}{0.32, 0.09, 0.98}
\definecolor{green(pigment)}{rgb}{0.0, 0.65, 0.31}

\makeatletter
\let\NAT@parse\undefined
\makeatother
\usepackage[pagebackref=false, breaklinks=true, colorlinks, bookmarks=false]{
        hyperref
}
\hypersetup{
        colorlinks=true,
        linkcolor={red},
        citecolor={hanpurple},
        urlcolor={magenta}
}

\usepackage{capt-of}
\usepackage{graphicx}

\title{\LARGE \bf
UniTrackPLA: Unified Panorama-Language-Action Model for Instruction-Guided Navigation and Dynamic Person Tracking}

\author{Pengfei Qi$^{1}$, Haoran Lin$^{2}$, Sizhuang Chen$^{2}$, Kai Luo$^{2}$, Sirui Zhang$^{2}$, Xinqi Liu$^{2}$, Fei Cheng$^{3,4}$,\\Wenrui Chen$^{2}$, Liming Yin$^{4}$, and Kailun Yang$^{1,2,\dag}$%
\thanks{This work was supported in part by the National Natural Science Foundation of China (Grant No. 62473139 and No. 62388101), in part by the Hunan Provincial Research and Development Project (Grant No. 2025QK3019), and in part by the State Key Laboratory of Autonomous Intelligent Unmanned Systems (the opening project number ZZKF2025-2-10).}%
\thanks{$^{1}$The authors are with the College of Integrated Circuits, Hunan University, Changsha, China (kailun.yang@hnu.edu.cn).}%
\thanks{$^{2}$The authors are with the School of Artificial Intelligence and Robotics and the National Engineering Research Center of Robot Visual Perception and Control Technology, Hunan University, Changsha, China.}%
\thanks{$^{3}$The author is with the School of Advanced Technology, Xi'an Jiaotong-Liverpool University, Suzhou, China.}%
\thanks{$^{4}$The authors are with Suzhou VSDeep Intelligent Technology Co., Ltd., Suzhou, China.}%
\thanks{$^{\dag}$Corresponding author: Kailun Yang.}
}

\let\oldtwocolumn\twocolumn
\renewcommand\twocolumn[1][]{%
    \oldtwocolumn[{#1}{
    \begin{center}
    \vskip -3ex
        \centering
        \includegraphics[width=\textwidth]{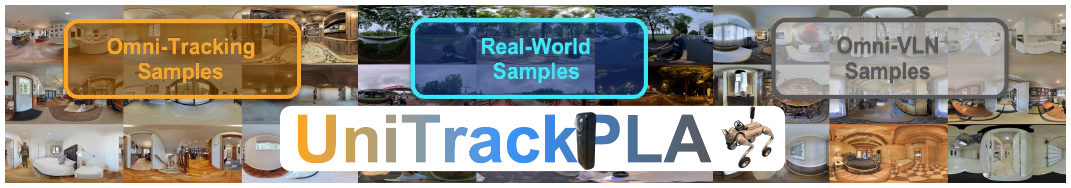}
        \vskip -1ex
        \captionof{figure}{{\textbf{UniTrackPLA} is a unified panorama-language-action model for instruction-guided navigation and dynamic person tracking. It is trained on Omni-Tracking, real-world panoramic, and Omni-VLN samples, enabling unified learning across simulated and real-world environments.
        }}
        \label{fig:teaser}
    \end{center}
    }]
}

\begin{document}

\maketitle
\thispagestyle{empty}
\pagestyle{empty}

\begin{abstract}
General-purpose embodied robots should support both navigation toward language-specified destinations and dynamic person tracking under arbitrary initial target azimuths. However, existing methods typically rely on forward-facing observations and address these tasks with separate policies, limiting omnidirectional perception and unified closed-loop control. We present \textbf{UniTrackPLA}, a unified panorama-language-action model for instruction-guided navigation and dynamic person tracking. Its Panoramic-Aware Encoding (PAE) preserves the temporal and azimuthal structure of perspective views projected from each panorama, enabling perspective-pretrained visual encoders to process omnidirectional observations. A shared vision-language backbone grounds instructions in the panoramic context and predicts continuous robot-centric waypoint chunks for both tasks. World-Action Consistency (WAC) further predicts action-conditioned future visual states and verifies waypoint prefixes online, allowing reliable actions to be reused while triggering replanning upon inconsistency. We also introduce \textbf{OmniTrackNav-Bench}, comprising $5,000$ simulated tracking trajectories, $10,000$ simulated VLN routes, and $96$ verified real-world routes, providing $919,978$ waypoint-supervision instances. UniTrackPLA improves overall tracking SR from $23.50\%$ to $35.00\%$ and Omni-VLN SR/SPL from $13.00\%$/$12.77\%$ to $19.75\%$/$19.29\%$. Incorporating $76$ real-world routes further improves held-out EP@0.2m from $42.92\%$ to $92.08\%$. Closed-loop experiments on a Go2-W robot demonstrate unified panoramic tracking and navigation across indoor and outdoor environments. The project page is at~\href{https://tw5775.github.io/UniTrackPLA}{https://tw5775.github.io/UniTrackPLA}.
\end{abstract}

\section{Introduction}
Embodied navigation enables robots to move through an environment according to visual observations and task instructions, constituting a fundamental capability for household assistance, human accompaniment, inspection, and mobile manipulation~\cite{wang2022towards,zhang2026embodied_foundation,cai2026navigation_wayfinding,yang2026learning_panorama}. Existing studies, however, have largely evolved along two separate paradigms. Vision-language navigation mainly focuses on reaching static semantic goals specified by language instructions~\cite{zhang2024uni_navid,qiao2025navbench,lu2025monovln,dai2026thinkmatter}, whereas embodied visual tracking emphasizes continuously following dynamic targets whose positions evolve over time~\cite{wang2025trackvla,wu2025vlm,sun2026instance,liu2026trackvla++,ye2026refertrack}. 
Recent Vision-Language-Action (VLA) models integrate visual perception, language understanding, and action prediction, improving instruction following and generalization in unseen environments~\cite{zhang2026embodied_foundation,zhang2024uni_navid,wang2025trackvla,li2026urbanvla}.
Nevertheless, static-goal navigation and dynamic person tracking are still commonly treated separately, with task-specific observations, action spaces, and policies.

We argue that these two tasks should instead be viewed as complementary instances of a broader embodied navigation problem. Instruction-guided navigation is primarily driven by relatively static semantic goals, while person tracking requires continuous adaptation to dynamically evolving physical targets. Despite this difference, both require the robot to ground task-relevant entities, reason about their spatial relationships, and continuously translate visual-language observations into motion decisions. 
A more general embodied navigation system should therefore be able to bridge these two regimes within a shared perception-language-action paradigm. This raises a fundamental question: \emph{can a single policy support both instruction-guided navigation toward static goals and continuous tracking of dynamic persons under a unified observation and action interface?}

Bridging static navigation and dynamic tracking requires a shared omnidirectional perceptual representation. 
Panoramic perception provides a natural means of removing this heading dependence by exposing the surrounding scene within a single omnidirectional observation space~\cite{zhang2026embodied_foundation,zhang2024uni_navid,luo2025omnitrack}.
Prior studies have explored panoramic RGB-D observations~\cite{watkins2020learning,kawaharazuka2024reflex}, omnidirectional images~\cite{wang2024enhanced}, multiple perspective views~\cite{lu2025monovln}, panoramic environment generation~\cite{li2023panogen,wang2025panogen++}, and latent panorama prediction~\cite{dai2026thinkmatter} for static-goal search and vision-language navigation. Recent approaches further incorporate panoramic scene parsing~\cite{jin2026panonav}, dynamic memory~\cite{sheng2026p2dnav}, and global-to-local reasoning~\cite{liu2026omnivln,damodaran2026eagor}. Nevertheless, existing panoramic systems mainly focus on static object search, route following, and scene reasoning, while dynamic target localization across arbitrary orientations and panorama boundaries remains insufficiently explored. More importantly, panoramic observations introduce a representation gap: most pretrained visual encoders are optimized for perspective images, whereas panoramic perception requires explicit reasoning over circular spatial structure and temporal evolution. Therefore, the key challenge is not merely to enlarge the field of view, but to retain the transferable representations of pretrained perspective-image encoders while recovering the spatial and temporal structure required for unified panoramic embodied navigation.

A unified policy must also handle the different temporal dynamics of static navigation and dynamic tracking. Continuous waypoint heads improve efficiency by predicting short-horizon action chunks~\cite{zhang2024uni_navid,wang2025trackvla}, but fixed-horizon execution may become unreliable as the visual world evolves: trajectories that remain valid in relatively stable navigation can quickly become stale when a tracked person changes direction, local traversability changes, or execution errors accumulate. Executing the full chunk risks propagating invalid actions, while using only the first waypoint sacrifices most of the predicted trajectory. The key challenge is therefore \emph{how long a predicted trajectory should remain trusted}. Since pixel-level world models are costly for real-time control~\cite{nie2025wmnav,yao2026sc2wm}, we instead employ lightweight action-conditioned latent future prediction to adaptively regulate the execution horizon.

To bridge static instruction-guided navigation and dynamic person tracking across both perception and execution, we propose \textbf{UniTrackPLA}, a unified Panorama-Language-Action framework. UniTrackPLA formulates both tasks under shared panoramic observations and a continuous waypoint action space. \textbf{Panoramic-Aware Encoding} (PAE) captures visual content, view azimuth, and temporal position while preserving compatibility with pretrained perspective-image encoders, and a shared multimodal policy predicts waypoint chunks for both navigation and tracking. We introduce \textbf{World-Action Consistency} (WAC), which predicts action-conditioned future states in latent space and adaptively determines whether to continue the current chunk or replan, enabling efficient yet responsive closed-loop control.

To support unified training and evaluation, we construct \textbf{OmniTrackNav-Bench}, comprising \textbf{Omni-Tracking} with $5,000$ dynamic-following trajectories and \textbf{Omni-VLN} with $10,000$ language-annotated panoramic navigation routes. Both tasks share the same panoramic observation and continuous waypoint representation, enabling unified multi-task learning within a single policy. We introduce a real-world
panoramic navigation dataset comprising $96$ routes.

Closed-loop experiments show that UniTrackPLA achieves a $35.00\%$ success rate on Omni-Tracking, compared with $23.50\%$ for our reproduced TrackVLA
baseline, and reaches $19.75\%$ SR and $19.29\%$ SPL on Omni-VLN, outperforming the baseline results of $13.00\%$ and $12.77\%$. Ablation studies verify the effectiveness of PAE and WAC. With this dataset, EP@0.2m
on held-out real-world routes improves from $42.92\%$ to $92.08\%$. Real-world deployment on a Go2-W robot with an Insta360 X4 further demonstrates both dynamic person tracking and instruction-guided navigation
using the same policy.

Our main contributions are summarized as follows:
\begin{itemize}
\item We propose \textbf{UniTrackPLA}, a panoramic VLA framework that bridges dynamic person following and vision-language navigation through shared panoramic perception, visual-language representations, and a continuous waypoint action space. Its PAE module jointly models panoramic azimuth and temporal context while reusing pretrained perspective-image encoders.

\item We introduce a lightweight \textbf{World-Action Consistency} module that unifies temporal action execution across static navigation and dynamic tracking. By predicting the latent future consequences of waypoint prefixes, WAC adaptively regulates the effective action horizon and dynamically determines whether to continue the current chunk or trigger replanning.

\item We construct the \textbf{OmniTrackNav-Bench} to support joint training and consistent evaluation across static navigation and dynamic tracking, comprising Omni-Tracking for arbitrary-orientation human following and Omni-VLN for dense panoramic language navigation. Extensive simulation experiments and real-robot deployments validate the effectiveness and generality of the overall framework.

\end{itemize}

\section{Related Work}

\subsection{Embodied Navigation}

Existing embodied navigation methods can be broadly divided into static-goal navigation and dynamic-person following. 
Static-goal navigation guides a robot toward a specified location according to language instructions, target categories, or reference routes. 
Recent methods unify multiple navigation tasks and improve generalization across environments, embodiments, and long-horizon routes~\cite{zhang2026embodied_foundation,zhang2024uni_navid,li2026urbanvla}, while observation-instruction augmentation further facilitates transfer to unseen environments~\cite{wei2026unseen_seen}.

\begin{figure*}[!t]
    \centering
    \includegraphics[width=\linewidth]{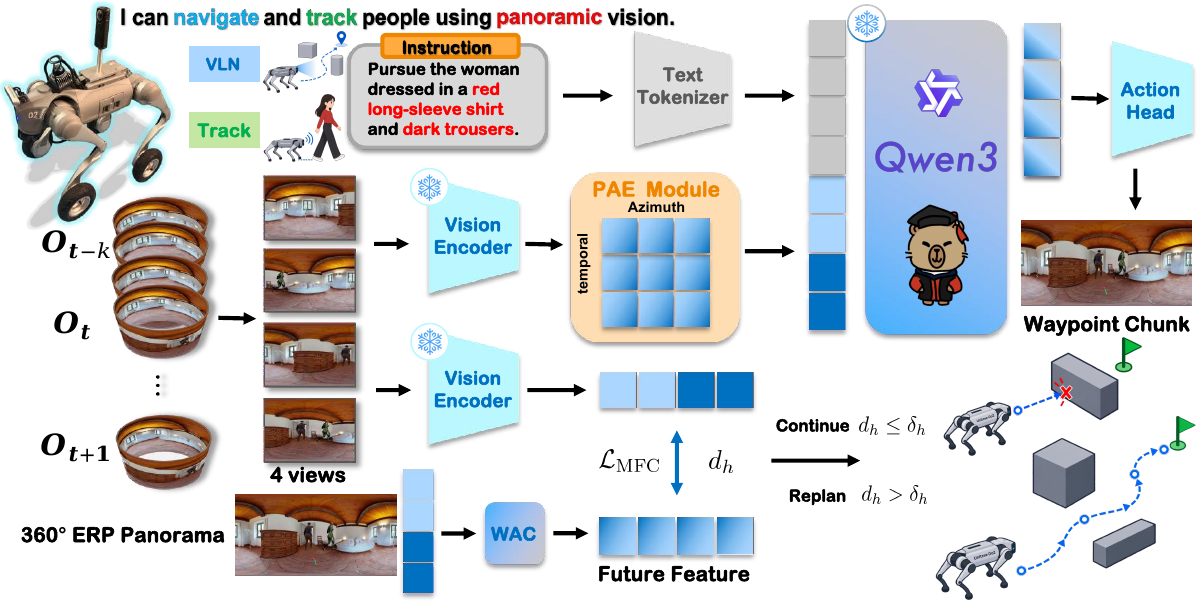}
    \vskip-1.5ex
    \caption{
    \textbf{Overview of the proposed UniTrackPLA pipeline.} 
    The upper policy branch encodes four perspective views with PAE and fuses them with language instructions to predict waypoint chunks. 
    The lower WAC branch predicts action-conditioned future features and evaluates future-state consistency to dynamically continue execution or trigger replanning. Snowflakes denote frozen modules.}
    \label{overall_pipeline}
    \vskip-4ex
\end{figure*}

Dynamic person following instead requires continuous target identification and motion adaptation. 
Hierarchical approaches bridge high-level user instructions and low-level tracking actions through intermediate spatial goals~\cite{wu2025hierarchical}. 
Other modular approaches employ visual foundation models, offline reinforcement learning, and vision-language models to improve target representation, failure recovery, occlusion handling, and cross-embodiment adaptation~\cite{wu2025vlm,sun2026instance,zhong2024empowering,wu2026adatracker}. End-to-end methods further incorporate referring comprehension, target memory, competitive interaction, and trajectory planning into vision-language-action policies~\cite{wang2025trackvla,liu2026trackvla++,ye2026refertrack,liu2026comatrack}, while recent benchmarks evaluate long-term tracking, socially aware following, and unified target seeking and following~\cite{ye2025tpt_bench,ye2025follow_bench,yu2026uesf}. However, most existing tracking methods rely on limited-FoV forward-facing cameras and assume that the target is initially visible in front of the robot, leaving side, rear, and panorama-boundary initialization underexplored.

\subsection{Panoramic-Aware Embodied Navigation}
Panoramic-aware navigation methods can be broadly divided into panoramic-perception-based navigation and panorama-assisted navigation.
The former directly uses panoramic RGB-D observations, omnidirectional images, or multiple perspective views for target-driven, open-vocabulary, and language-guided navigation~\cite{watkins2020learning,kawaharazuka2024reflex,wang2024enhanced}.
In particular, view-decomposed panoramic representations preserve local visual details while retaining compatibility with visual encoders pretrained on perspective images. 
The latter leverages generated panoramic environments, complementary views, or predicted latent panoramic features to bridge the observation gap between monocular and panoramic navigation~\cite{lu2025monovln,dai2026thinkmatter,li2023panogen,wang2025panogen++}.
Recent methods further combine panoramic scene parsing, dynamic memory, panorama-to-local reasoning, and 3D scene representations for mapless object navigation and zero-shot vision-language navigation~\cite{jin2026panonav,sheng2026p2dnav,liu2026omnivln}, while other studies investigate spherical directional reasoning and establish benchmarks for panoramic urban navigation and general embodied navigation~\cite{qiao2025navbench,damodaran2026eagor,watanabe2026_360cityarena}.
Beyond navigation, omnidirectional perception has also enabled beyond-FoV humanoid manipulation and panorama-aware mobile manipulation~\cite{yang2026learning_panorama,qu2026omnidp}. 
Nevertheless, existing panoramic embodied systems primarily focus on static object search, route following, scene reasoning, or manipulation, leaving dynamic person following under arbitrary initial orientations, particularly cross-view localization and identity preservation, less explored. 
Moreover, navigation and person following are commonly modeled using separate representations, action spaces, and policies. 
A unified panoramic vision-language-action policy for both instruction-guided navigation and dynamic person following remains insufficiently studied.

\section{Method}

\textbf{UniTrackPLA Overview.}
In this section, we put forward the Unified Panorama-Language-Action (UniTrackPLA) framework to unify language-guided navigation and dynamic-person following by leveraging panoramic observations within a single embodied policy. 
An overview of the proposed UniTrackPLA model is shown in Fig.~\ref{overall_pipeline}. 
First, the \textbf{Panoramic-Aware Encoding} (PAE) module (see Sec.~\ref{sec:pae}) projects each panorama into cyclic perspective views and preserves their temporal and azimuthal identities, adapting perspective-pretrained visual representations to panoramic observations.
Next, a shared vision-language policy (see Sec.~\ref{sec:unified_policy}) grounds task instructions in the panoramic context and predicts a chunk of robot-centric waypoints for both following and navigation. Finally, the \textbf{World-Action Consistency} (WAC) module (see Sec.~\ref{sec:wac}) predicts the future visual consequences of waypoint prefixes and verifies their consistency during execution, enabling the controller to reuse reliable actions or trigger replanning when the environment deviates from the predicted transition.

\subsection{Panoramic-Aware Encoding}
\label{sec:pae}

Visual encoders pretrained on perspective images are suboptimal for panoramic
observations. Directly encoding an equirectangular projection (ERP) causes
non-uniform distortion, while independently encoding perspective crops loses
their spatial arrangement, hindering cross-view target association and
directional grounding.

We introduce \textbf{Panoramic-Aware Encoding} (PAE) to preserve the azimuthal and temporal identities of panoramic features. Each ERP is projected into $N$ cyclically ordered perspective views with a configurable
field of view. Frozen DINOv3~\cite{simeoni2025dinov3} and
SigLIP~\cite{zhai2023sigmoid} encoders extract complementary features, which are concatenated and mapped into the language-model embedding space by a trainable projector.

For token $j$ at azimuth slot $m$ and time $t$, PAE computes
\begin{equation}
\widetilde{\mathbf{v}}_{t,m,j}
=
\mathbf{P}\mathbf{v}_{t,m,j}
+
\mathbf{e}^{\mathrm{azi}}_{m},
\label{eq:pae}
\end{equation}
where $\mathbf{P}$ is the multimodal projector and
$\mathbf{e}^{\mathrm{azi}}_{m}$ identifies the panoramic direction. A temporal
token $\mathbf{e}^{\mathrm{temp}}_{t}+\mathbf{e}^{\mathrm{type}}_{t}$ is
prepended to each frame block to encode its time step and distinguish
historical from current observations. Frame blocks are ordered temporally,
with views cyclically ordered within each block.

For efficiency, fine-grained patch tokens are retained only for the current
views, while each historical view is represented by a compact global token.
This preserves the temporal-azimuthal structure required for panoramic
tracking and navigation with limited context overhead.

\subsection{Unified Vision-Language-Action Policy}
\label{sec:unified_policy}

Language identifies a target by appearance in person following, while
describing routes, landmarks, and destinations in VLN. Despite these
different semantics, both tasks require continuous local motion planning from
panoramic observations. We therefore use a unified vision-language-action
policy that shares its panoramic representation, multimodal backbone, and
action space, with task behavior determined by the instruction and visual
context.

The panoramic tokens from PAE and the instruction tokens are jointly
processed by a frozen Qwen3-0.6B backbone~\cite{qwen3}. The language backbone
is frozen to preserve its pretrained semantics, while the multimodal
projector, PAE embeddings, and continuous action head remain trainable.

Rather than generating discretized action tokens autoregressively, the action
head directly predicts a waypoint chunk of length $K=8$:
\begin{equation}
\widehat{\mathcal A}_t
=
f_{\mathrm{act}}(\mathbf h_t)
=
\left\{
(\Delta x_t^k,\Delta y_t^k,\Delta\omega_t^k)
\right\}_{k=1}^{K},
\label{eq:waypoint_policy}
\end{equation}
where $\mathbf h_t$ is the multimodal state representation. Each tuple
specifies the cumulative planar pose of waypoint $k$ relative to the current
robot frame, rather than an increment from waypoint $k-1$. The resulting
chunk captures short-horizon motion and conditions WAC for future-feature
prediction and consistency verification.

The policy follows the language-referred target and its cross-view motion in
person following, while predicting local paths from route instructions and
observed landmarks in VLN. Both tasks therefore share a single panoramic
policy and continuous waypoint representation.

\subsection{World-Action Consistency}
\label{sec:wac}

Waypoint chunking provides smooth short-horizon plans, but behavior cloning
does not model the future visual states induced by predicted actions.
Consequently, later waypoints may become invalid under target motion,
environmental changes, or execution errors. We introduce
\textbf{World-Action Consistency} (WAC) to predict compact future visual
features conditioned on the current state and waypoint prefixes, without
reconstructing RGB observations.

At time $t$, the current fine-grained panoramic tokens are mean-pooled and
$\ell_2$-normalized into a visual state $\mathbf z_t$. Each waypoint is
projected into an action token, and a future query is assigned to each
prediction horizon. An action-conditioned Transformer predicts
\begin{equation}
\widehat{\mathbf z}_{t+h}
=
\mathcal W_{\theta}
\left(
\mathbf h_t,\,
\mathbf z_t,\,
\mathcal A_t^{1:h}
\right),
\qquad
h=1,\ldots,H_w,
\label{eq:wac_prediction}
\end{equation}
where $\mathbf h_t$ is the multimodal policy state and
$\mathcal A_t^{1:h}$ denotes the first $h$ waypoints. The policy predicts
$K=8$ waypoints, while WAC models the first $H_w=4$ prefixes.

The Transformer interleaves action tokens with horizon-specific queries. Its
causal mask restricts each query to the current state and its corresponding
action prefix, preventing information leakage from later waypoints.

During training, WAC is conditioned on expert waypoint chunks, and target
states are obtained by encoding, pooling, and normalizing future panoramas.
The policy states, visual states, actions, and targets are detached so that
the auxiliary objectives update only WAC. At inference, WAC instead receives
the policy-predicted chunk. An inverse-action head additionally reconstructs
the expert chunk from the current and longest-horizon states, encouraging the
predicted transition to remain action-sensitive. All predictions operate in
latent space, avoiding pixel-level generation.

\subsection{Consistency-Guided Closed-Loop Execution}
\label{sec:closed_loop_execution}

Multi-step prediction improves planning efficiency, but a fixed execution
horizon creates a trade-off: executing the entire chunk may retain invalid
actions after a state change, whereas executing only one waypoint requires
frequent policy inference. We therefore use WAC to determine online how much
of each waypoint chunk remains valid.

Given the panoramic history and instruction, the policy predicts $K=8$
waypoints, while WAC estimates the future states for the first $H_w=4$
prefixes. The robot first executes the first waypoint. After executing the $h$-th waypoint, it encodes the resulting panorama as $\mathbf z^{\mathrm{obs}}_{t+h}$. Its discrepancy from the predicted state is
\begin{equation}
d_h
=
1-
\operatorname{cos}
\left(
\widehat{\mathbf z}_{t+h},
\mathbf z^{\mathrm{obs}}_{t+h}
\right).
\label{eq:wac_discrepancy}
\end{equation}
A small $d_h$ indicates that execution remains consistent with the predicted
transition, while a large value indicates a change in the target, environment,
or robot state.

For $h < H_w$, the controller applies a horizon-specific threshold
$\delta_h$:
\begin{equation}
\begin{cases}
\text{Continue}, & d_h \leq \delta_h,\\
\text{Replan},   & d_h > \delta_h.
\end{cases}
\label{eq:wac_decision}
\end{equation}
If consistent, the next waypoint is executed without another policy query. Otherwise, the remaining suffix is discarded, and a new chunk is generated from the latest observation and updated history. Replanning starts directly from the current robot state.

Because WAC verifies only the first $H_w$ horizons, at most $H_w$ waypoints are executed from each chunk before the policy generates a new one, even when all transitions remain consistent. This mechanism turns each fixed-length chunk into an online-verifiable plan, balancing trajectory reuse with responsive closed-loop control.

\subsection{Optimization Objective}
\label{sec:optimization}
We jointly optimize waypoint regression, multi-horizon future-state prediction, and inverse-action reconstruction. The waypoint loss $\mathcal L_{\mathrm{wp}}$ is a masked mean-squared error between predicted and expert waypoint chunks in the normalized action space, excluding unavailable future waypoints.

WAC is supervised by the cosine distance between predicted and target future features:
\begin{equation}
\mathcal L_{\mathrm{mfc}}
=
\frac{
\sum_{h=1}^{H_w}
m_h
\left[
1-
\operatorname{cos}
\left(
\widehat{\mathbf z}_{t+h},
\mathbf z_{t+h}
\right)
\right]
}{
\sum_{h=1}^{H_w}m_h
},
\label{eq:mfc_loss}
\end{equation}
where $m_h$ indicates the availability of the target at horizon $h$, and both features are $\ell_2$-normalized. The inverse-action loss $\mathcal L_{\mathrm{inv}}$ is a masked mean-squared error for reconstructing the normalized expert waypoint chunk. It prevents WAC from predicting visually plausible future states unrelated to the conditioned actions. The complete objective is
\begin{equation}
\mathcal L
=
\lambda_{\mathrm{wp}}\mathcal L_{\mathrm{wp}}
+
\lambda_{\mathrm{mfc}}\mathcal L_{\mathrm{mfc}}
+
\lambda_{\mathrm{inv}}\mathcal L_{\mathrm{inv}},
\label{eq:total_objective}
\end{equation}
with weights specified in the implementation details. Inputs and targets of
the WAC objectives are detached, restricting auxiliary gradients to WAC. At
inference, the future targets and inverse-action head are removed, while the
WAC predictor is retained for consistency-guided execution.

\section{Experiment Setup}

\subsection{Dataset}
\label{sec:dataset}
\begin{table*}[!t]
\centering
\caption{Quantitative comparison between TrackVLA and UniTrackPLA on the Omni-Tracking evaluation split.}
\label{tab:main_tracking_comparison}
\vskip-1ex
\setlength{\tabcolsep}{4.2pt}
\renewcommand{\arraystretch}{1.10}
\begin{tabular}{lcccccccccccc}
\toprule
\multirow{2}{*}{Method}
& \multicolumn{4}{c}{STF}
& \multicolumn{4}{c}{DRF}
& \multicolumn{4}{c}{Overall} \\
\cmidrule(lr){2-5}
\cmidrule(lr){6-9}
\cmidrule(lr){10-13}
& SR$\uparrow$ & FR$\uparrow$ & LR$\downarrow$ & CR$\downarrow$
& SR$\uparrow$ & FR$\uparrow$ & LR$\downarrow$ & CR$\downarrow$
& SR$\uparrow$ & FR$\uparrow$ & LR$\downarrow$ & CR$\downarrow$ \\
\midrule

TrackVLA$^\dagger$
& 35.00 & 65.35 & 27.00 & 20.00
& 12.00 & 56.27 & 48.00 & 20.00
& 23.50 & 60.81 & 37.50 & 20.00 \\

\rowcolor{gray!20}
\textbf{Ours}
& \textbf{51.00} & \textbf{75.65} & \textbf{14.00} & \textbf{15.00}
& \textbf{19.00} & \textbf{58.83} & \textbf{46.00} & \textbf{16.00}
& \textbf{35.00} & \textbf{67.24} & \textbf{30.00} & \textbf{15.50} \\
\bottomrule
\end{tabular}

%\vspace{2pt}
\parbox{\textwidth}{\footnotesize
$^\dagger$TrackVLA denotes our reproduced baseline evaluated under the same panoramic observation and closed-loop evaluation protocol.}
\vskip-2ex
\end{table*}

\begin{table*}[!t]
\centering
\caption{Quantitative comparison between TrackVLA and Ours on the Omni-VLN evaluation split.}
\label{tab:main_navigation_comparison}
\vskip-1ex
\setlength{\tabcolsep}{5.0pt}
\renewcommand{\arraystretch}{1.10}
\begin{tabular}{lcccccccccc}
\toprule
\multirow{2}{*}{Method}
& \multicolumn{2}{c}{R2R-CE Seen}
& \multicolumn{2}{c}{R2R-CE Unseen}
& \multicolumn{2}{c}{RxR-CE Seen}
& \multicolumn{2}{c}{RxR-CE Unseen}
& \multicolumn{2}{c}{Overall} \\
\cmidrule(lr){2-3}
\cmidrule(lr){4-5}
\cmidrule(lr){6-7}
\cmidrule(lr){8-9}
\cmidrule(lr){10-11}
& SR$\uparrow$ & SPL$\uparrow$
& SR$\uparrow$ & SPL$\uparrow$
& SR$\uparrow$ & SPL$\uparrow$
& SR$\uparrow$ & SPL$\uparrow$
& SR$\uparrow$ & SPL$\uparrow$ \\
\midrule

TrackVLA$^\dagger$
& 15.00 & 15.00
& 14.00 & 13.80
& 11.00 & 10.62
& 12.00 & 11.66
& 13.00 & 12.77 \\

\rowcolor{gray!20}
\textbf{Ours}
& \textbf{19.00} & \textbf{18.40}
& \textbf{20.00} & \textbf{20.00}
& \textbf{19.00} & \textbf{17.93}
& \textbf{21.00} & \textbf{20.88}
& \textbf{19.75} & \textbf{19.30} \\
\bottomrule
\end{tabular}

%\vspace{2pt}
\parbox{\textwidth}{\footnotesize
$^\dagger$TrackVLA denotes our reproduced baseline evaluated using the same panoramic observations and closed-loop evaluation protocol.}
\vskip-3ex
\end{table*}
Existing embodied navigation benchmarks commonly assume a limited forward-facing field of view and treat person following and vision-language navigation as separate tasks. 
We introduce \textbf{OmniTrackNav-Bench}, a unified benchmark for both tasks under omnidirectional perception. 
It comprises \textbf{Omni-Tracking}, \textbf{Omni-VLN}, and a \textbf{Real-World Dataset}, containing $5,000$ simulated person-following trajectories, $10,000$ simulated VLN routes, and $96$ verified real-world routes. 
Together, they provide $919,978$ waypoint-supervision instances using $360^{\circ}$ ERP observations and a shared continuous waypoint space.

\begin{figure}[!t]
    \centering
    \includegraphics[width=\columnwidth]{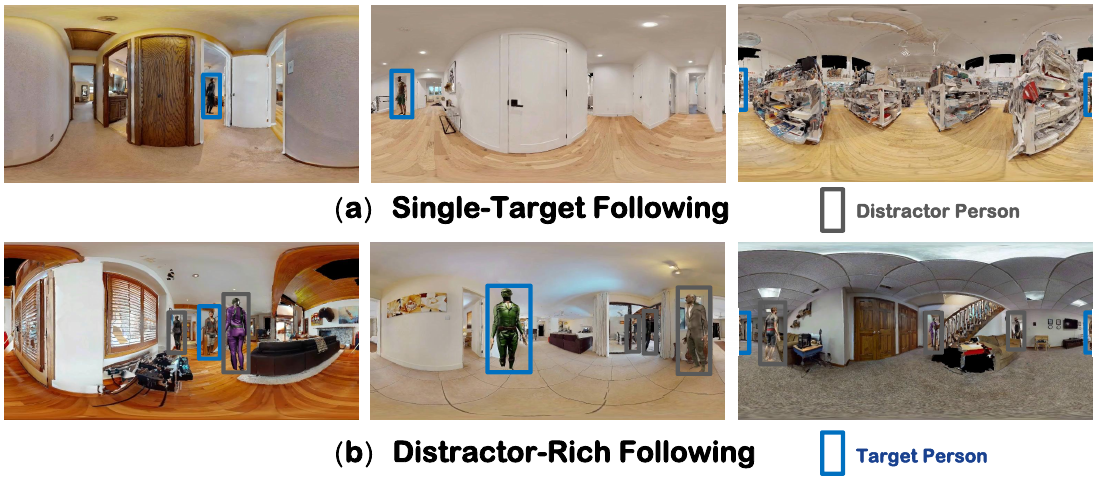}
    \vskip-1.5ex
    \caption{\textbf{Overview of the Omni-Tracking dataset.}
    It contains single-target following and distractor-rich following under arbitrary $360^{\circ}$ target initialization. Blue and gray boxes denote the referred target and distractor persons, respectively.}
    \label{fig:overview_track_dataset}
    \vskip-3ex
\end{figure}

\noindent\textbf{Omni-Tracking.}
As shown in Fig.~\ref{fig:overview_track_dataset}, Omni-Tracking contains $5,000$ panoramic person-following trajectories and $399,294$ supervision instances. The trajectories are evenly divided into \emph{Single-Target Following} (STF) and \emph{Distractor-Rich Following} (DRF), with $2,500$ trajectories per setting. Each episode initializes the referred person at an arbitrary azimuth, allowing the target to appear in front of, beside, behind, or across the ERP seam. STF evaluates target localization and persistent following, whereas DRF introduces multiple people and requires language-based grounding and identity maintenance.

Unlike forward-view tracking datasets that generate trajectories from image-space cues such as bounding-box size and pixel count, these cues are unstable in ERP panoramas because target appearance varies with azimuth, distance, and seam location. We therefore generate trajectories from the simulator-provided robot--target relative position, distance, and heading, subject to explicit following-distance constraints. Omni-Tracking further contains a validation split of $60$ episodes, including $30$ STF and $30$ DRF episodes, and an evaluation split of $200$ episodes, including $100$ per setting. The training, validation, and evaluation splits have no episode or source-route overlap.

\noindent\textbf{Omni-VLN.}
Omni-VLN is constructed from R2R-CE and RxR-CE, each contributing $5,000$ training routes. Since the original routes provide observations only at sparse navigation nodes, we densely resample the executable paths and render ERP observations at the sampled poses, producing continuous panoramic navigation sequences.

At each decision point, a local future trajectory is represented as a fixed-length waypoint chunk. Each waypoint specifies cumulative planar displacement and yaw change in the current robot frame. The R2R-CE and RxR-CE
subsets contain $196,136$ and $288,281$ supervision instances, respectively, totaling $484,417$ navigation decisions. The evaluation split contains $400$ routes, evenly distributed across R2R-CE Seen, R2R-CE Unseen, RxR-CE Seen, and RxR-CE Unseen.

\begin{figure}[!t]
    \centering
    \includegraphics[width=\columnwidth]{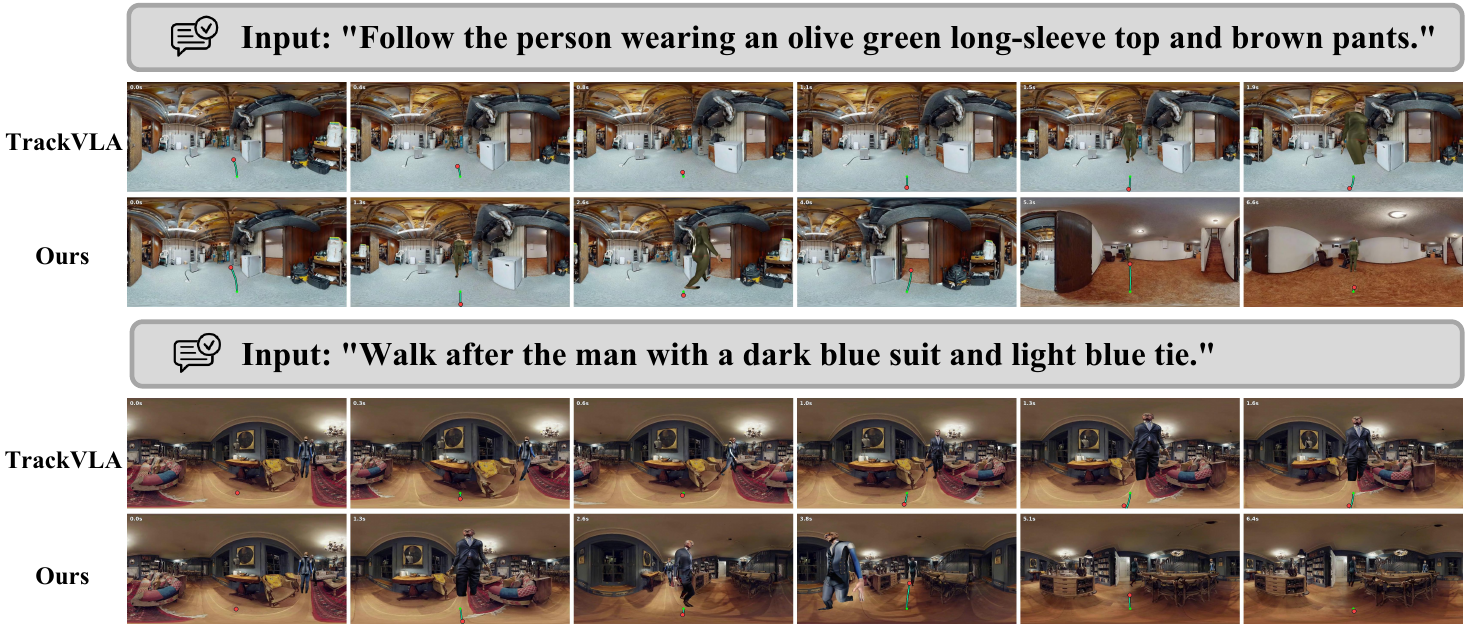}
    \vskip-1ex
    \caption{\textbf{Qualitative comparison between TrackVLA and Ours
for closed-loop panoramic person following.} The top and bottom examples
correspond to STF and DRF, respectively.}
    \label{fig:qualitative_tracking}
    \vskip-3ex
\end{figure}

\noindent\textbf{Real-World Dataset.}
We collect long-range trajectories using a handheld panoramic camera in
diverse indoor and outdoor environments, including roads, intersections,
pedestrian paths, plazas, parking areas, building entrances, lakeside
walkways, bridges, commercial streets, shopping malls, supermarkets, and offices. 
The data cover indoor-outdoor transitions and both daytime and nighttime conditions. 
Each $360^\circ$ ERP video is synchronized with its metric pose trajectory.

After temporal alignment and quality control, trajectories are divided into local navigation segments. At each step, future positions are interpolated at path-distance offsets of $0.1$--$0.8\,\mathrm{m}$ and transformed into the local frame, yielding eight cumulative waypoints $(\Delta x,\Delta y,\Delta\omega)$. Their headings are derived from smoothed trajectory tangents.

The dataset contains $96$ verified routes covering $9.13\,\mathrm{km}$, with average, median, and maximum lengths of $95.1\,\mathrm{m}$, $102.3\,\mathrm{m}$, and $120\,\mathrm{m}$, respectively. The routes yield $924$ local segments and $36,267$ waypoint-supervision samples. 
We use $76$ routes for training and $20$ route-disjoint routes for held-out evaluation, with nested $10$-, $50$-, and $76$-route training subsets for studying data scaling.

For instruction annotation, ERP keyframes are projected into perspective
views to identify static landmarks. Qwen3.6-27B uses ordered observations and
landmark evidence to generate and verify local navigation instructions. It is
used only for visual grounding and language annotation; all waypoint labels
are derived from measured trajectories using the same action representation
as Omni-VLN.

\subsection{Implementation Details}
\label{sec:implementation_details}

All models are trained for one epoch on eight NVIDIA RTX 3090 GPUs. DINOv3, SigLIP, and Qwen3-0.6B are frozen, while the multimodal projector, PAE, action head, and WAC are optimized. 
Based on the validation results, which will unfold in ablation analyses, each panorama is projected into four overlapping views at a resolution of $384 \times 384$, each with a horizontal field of view of $180^{\circ}$.
The policy uses up to $31$ historical observations and predicts eight waypoints, of which the first four are modeled by WAC.
We use AdamW with learning rates of $4\times10^{-5}$ for the main policy and $1\times10^{-4}$ for WAC, a weight decay of $0.01$, and gradient clipping at $1.0$. The per-GPU batch size is $8$, yielding a global batch size of $64$, and training uses bfloat16 precision. The weights of the waypoint, future-consistency, and inverse-action losses are $10$, $0.1$, and $0.05$, respectively. 
All hyperparameters are selected on the validation sets and fixed for final evaluation.

\noindent\textbf{Evaluation Metrics.}
SR measures success; FR, LR, and CR denote following, target-loss, and
collision rates. SPL reflects path efficiency. ADE, FDE, and Yaw MAE measure
average, final, and yaw errors, while EP@0.2m measures endpoint accuracy within
$0.2\,\mathrm{m}$. Arrows indicate preference.

\begin{table}[!t]
\centering
\caption{Effect of real-world navigation data on held-out route performance.}
\label{tab:real_data_scale}
\vskip-1ex
\setlength{\tabcolsep}{3.5pt}
\renewcommand{\arraystretch}{1.05}
\begin{tabular}{lcccc}
\toprule
Training Data
& ADE$\downarrow$
& FDE$\downarrow$
& Yaw MAE$\downarrow$
& EP@0.2m$\uparrow$ \\
\midrule
Sim-only
& 0.172
& 0.288
& 12.78
& 42.92 \\

Sim + 10 Real Routes
& 0.046 & 0.085 & 6.05 & 90.00 \\

Sim + 50 Real Routes
& 0.042
& 0.078
& 5.19
& 90.83\\

\rowcolor{gray!20}
\textbf{Sim + 76 Real Routes}
& \textbf{0.038}
& \textbf{0.069}
& \textbf{5.11}
& \textbf{92.08} \\

\bottomrule
\end{tabular}

%\vspace{1pt}
\vskip-3ex
\end{table}

\subsection{Quantitative Results}

\noindent\textbf{Human Tracking Results.}
Table~\ref{tab:main_tracking_comparison} shows that UniTrackPLA outperforms TrackVLA across all metrics. 
It improves overall SR and FR by $11.5$ and $6.4$ percentage points, respectively, while reducing loss and collision rates.
In the STF (top) and DRF (bottom) examples of Fig.~\ref{fig:qualitative_tracking}, TrackVLA approaches the targets aggressively and terminates after only $1.9s$ and $1.6s$, whereas UniTrackPLA maintains safer, stable following for $6.6s$ and $6.4s$ across scene transitions.

\noindent\textbf{Vision-Language Navigation Results.}
Table~\ref{tab:main_navigation_comparison} shows that UniTrackPLA consistently outperforms TrackVLA across all splits, with the largest gains in unseen environments.
Overall, it improves SR from $13.00\%$ to $19.75\%$ and SPL from $12.77\%$ to $19.29\%$, demonstrating better navigation success without sacrificing path efficiency.

\noindent\textbf{Effects of Incorporating Real-World Panoramic Navigation Data.}
Table~\ref{tab:real_data_scale} shows that only $10$ real routes reduce ADE from $0.172$ to $0.046$ and FDE from $0.288$ to $0.085$, while increasing EP@0.2m from $42.92\%$ to $90.00\%$. Using $76$ routes further achieves the best results of $0.038$ ADE, $0.069$ FDE, and $92.08\%$ EP@0.2m, demonstrating the benefit of real panoramic data for sim-to-real transfer. 
This shows UniTrackPLA effectively unlocks the potential of panoramic data to
improve real-world navigation performance.

\subsection{Ablation Studies}
We evaluate the panoramic view configuration and the individual contributions of PAE and WAC.

\subsubsection{Panoramic View Configuration}
Table~\ref{tab:panoramic_configuration} shows that four $180^\circ$ views provide the best balance, achieving the highest SR and FR and the lowest LR. Compared with a single $360^\circ$ view, this configuration improves SR by $8.34$ percentage points and reduces LR by $10.00$ points; we therefore adopt it by default. Narrower FoVs provide insufficient overlap, whereas additional views introduce redundant information.

\subsubsection{Effectiveness of PAE and WAC}
Table~\ref{tab:module_ablation} shows that PAE alone improves overall SR from
$23.50\%$ to $32.50\%$. WAC alone raises overall SR to $28.50\%$ but reduces
DRF SR from $12.00\%$ to $10.00\%$, indicating that consistency verification
benefits from explicit panoramic directional modeling. Combining PAE and WAC
achieves the best overall performance, confirming their complementary roles
in panoramic representation and closed-loop execution verification.
\begin{table}[!t]
\centering
\caption{Comparison of panoramic view configurations on the direction-balanced configuration split.}
\label{tab:panoramic_configuration}
\vskip-1ex
\setlength{\tabcolsep}{6pt}
\renewcommand{\arraystretch}{1.08}
\begin{tabular}{lcccc}
\toprule
Configuration
& SR$\uparrow$
& FR$\uparrow$
& LR$\downarrow$
& CR$\downarrow$ \\
\midrule

$N=1,\ \mathrm{FoV}=360^\circ$
& 13.33
& 59.22
& 31.67
& 48.33 \\

$N=2,\ \mathrm{FoV}=180^\circ$
& 5.00
& 56.81
& 36.67
& 53.33 \\

$N=4,\ \mathrm{FoV}=90^\circ$
& 13.33
& 55.82
& 33.33
& 41.67 \\

\rowcolor{gray!20}
$\mathbf{N=4,\ \mathrm{FoV}=180^\circ}$
& \textbf{21.67}
& \textbf{61.90}
& \textbf{21.67}
& 45.00 \\

$N=6,\ \mathrm{FoV}=60^\circ$
& 2.60 & 48.96 & 58.33 & \textbf{28.13} \\

$N=6,\ \mathrm{FoV}=120^\circ$
& 3.33 & 15.15 & 41.67 & 46.67 \\

\bottomrule
\end{tabular}
\vskip-2ex
\end{table}

\begin{table}[!t]
\centering
\caption{Ablation study of PAE and WAC.}
\label{tab:module_ablation}
\vskip-1ex
\setlength{\tabcolsep}{5.0pt}
\renewcommand{\arraystretch}{1.10}
\begin{tabular}{cc|cc|cccc}
\toprule
\multicolumn{2}{c|}{Components}
& \multicolumn{2}{c|}{Task SR$\uparrow$}
& \multicolumn{4}{c}{Overall} \\
\cmidrule(lr){1-2}
\cmidrule(lr){3-4}
\cmidrule(lr){5-8}
PAE & WAC
& STF & DRF
& SR$\uparrow$ & FR$\uparrow$ & LR$\downarrow$ & CR$\downarrow$ \\
\midrule

\ding{55} & \ding{55}
& 35.00 & 12.00
& 23.50 & 60.81 & 37.50 & 20.00 \\

\ding{51} & \ding{55}
& 47.00 & 18.00
& 32.50 & 67.23 & 32.00 & 16.00 \\

\ding{55} & \ding{51}
& 47.00 & 10.00
& 28.50 & 62.23 & 32.00 & 19.00 \\

\rowcolor{gray!20}
\ding{51} & \ding{51}
& \textbf{51.00} & \textbf{19.00}
& \textbf{35.00} & \textbf{67.24}
& \textbf{30.00} & \textbf{15.50} \\
\bottomrule
\end{tabular}
\vskip-3ex
\end{table}

\begin{figure}[!t]
    \centering
    \includegraphics[width=\columnwidth]{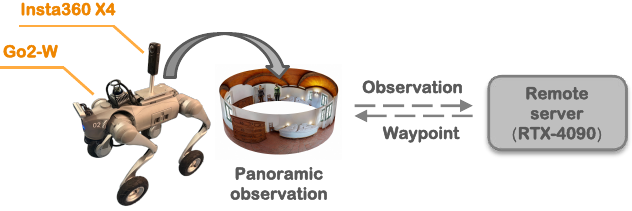}
    \vskip-1.5ex
    \caption{\textbf{Real-world system architecture.}
    An Insta360 X4 captures panoramic observations on the Go2-W robot, while a remote RTX 4090 server performs inference and returns waypoint commands.}
    \label{fig:Deployment}
    \vskip-2ex
\end{figure}

\begin{figure}[!t]
    \centering
    \includegraphics[width=\columnwidth]{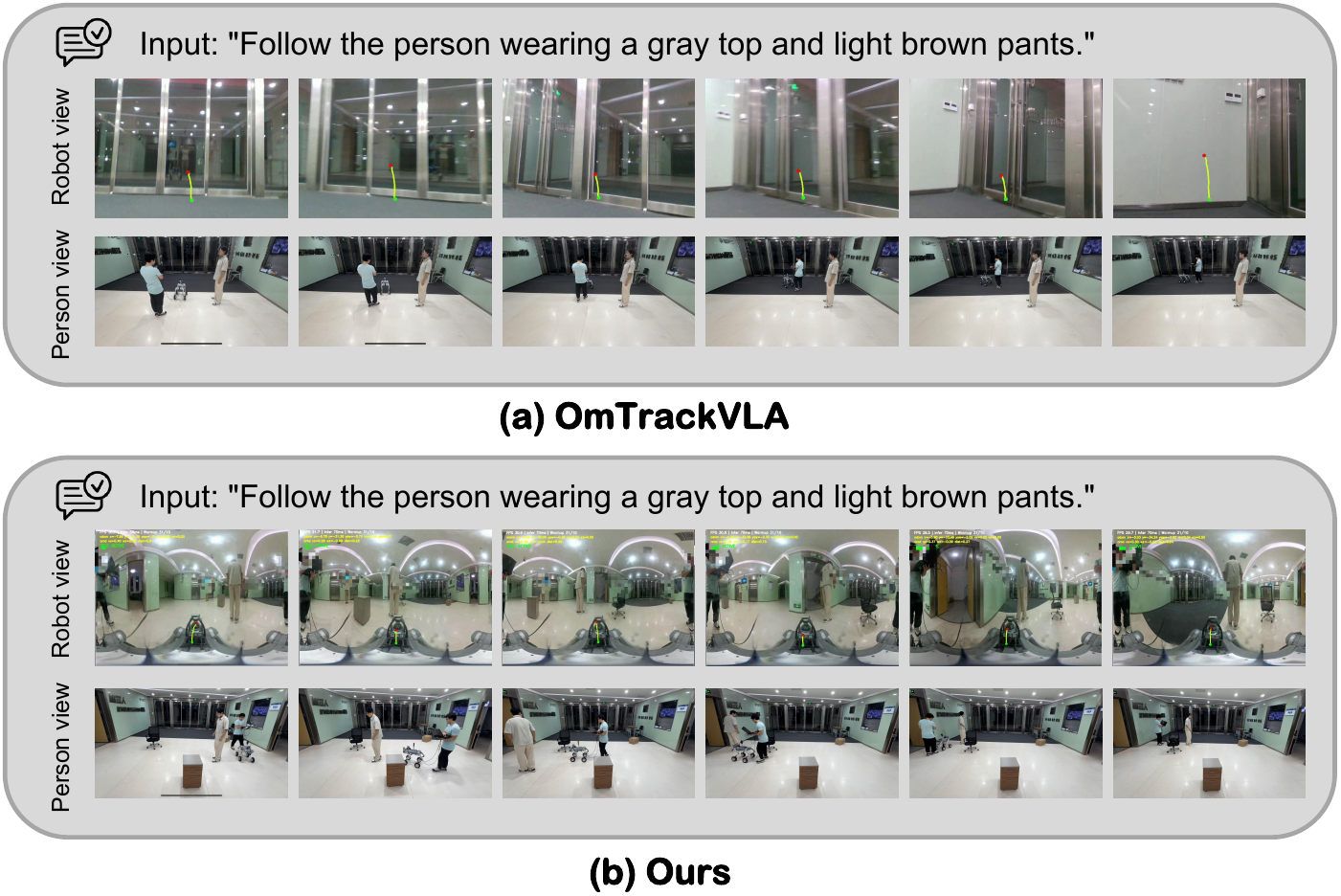}
    \vskip-1.5ex
    \caption{\textbf{Real-world person-following comparison in an office.}
    TrackVLA~\cite{wang2025trackvla} uses forward-facing RGB observations, whereas UniTrackPLA uses $360^{\circ}$ panoramic observations.}
    \label{fig:real_track_compare}
    \vskip-2ex
\end{figure}

\begin{figure}[!t]
    \centering
    \includegraphics[width=\columnwidth]{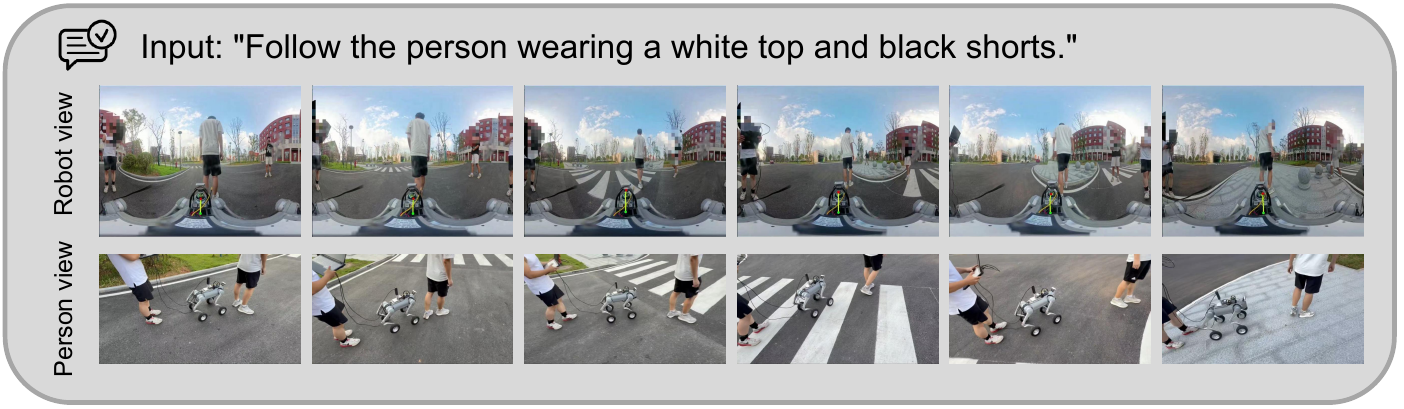}
    \vskip-1.5ex
    \caption{\textbf{Real-world outdoor person following.}}
    \label{fig:real_world_deployment_outdoor}
    \vskip-2ex
\end{figure}

\begin{figure}[!t]
    \centering
    \includegraphics[width=\columnwidth]{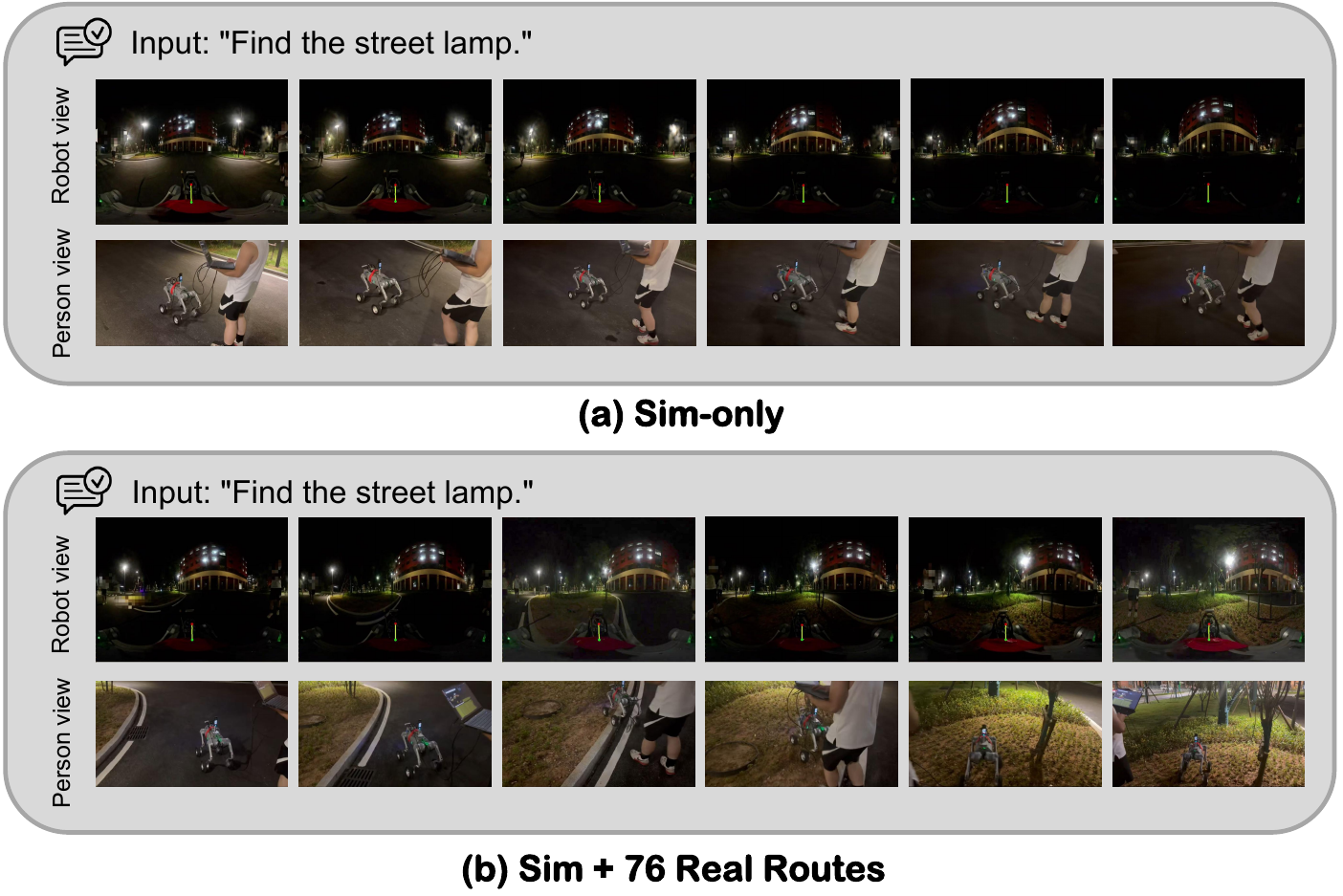}
    \vskip-1.5ex
    \caption{\textbf{Effect of real-world training data on outdoor navigation.}
    (a) Navigation using Sim-only training and (b) navigation using Sim + $76$ Real routes.}
    \label{fig:real_world_outdoor_navigation_ablation}
    \vskip-3ex
\end{figure}

\begin{figure*}[!t]
    \centering
    \includegraphics[width=\linewidth]{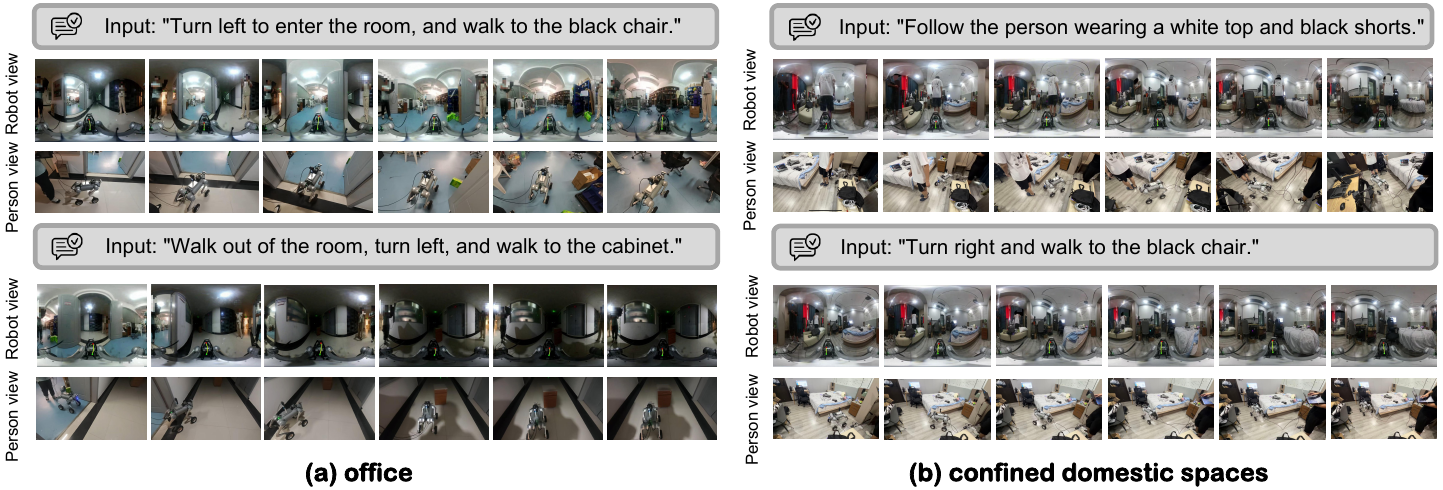}
    \vskip-1.5ex
    \caption{\textbf{Real-world experiments in diverse indoor environments.}
(a) Panoramic navigation in an office and (b) navigation and person tracking in confined domestic spaces.}
    \label{fig:real_world_office_domestic_spaces}
    \vskip-3ex
\end{figure*}

\subsection{Real-World Deployment Results}
As shown in Fig.~\ref{fig:Deployment}, the panoramic camera is connected to a local computer, which communicates with the remote inference server over WiFi. The Insta360 X4 captures $2880\times1440$ panoramic images. On a single NVIDIA RTX 4090 GPU, UniTrackPLA runs at $6.9$ FPS, while the robot executes closed-loop motion commands at $3.5$ Hz, allowing model inference to keep pace with physical execution. This design offloads model inference while retaining stable onboard observation acquisition.

Fig.~\ref{fig:real_track_compare} shows that the forward-view RGB TrackVLA~\cite{wang2025trackvla} loses the target when the person moves to the robot's side, whereas UniTrackPLA maintains tracking through panoramic perception. Further deployments demonstrate panoramic navigation in an office (Fig.~\ref{fig:real_world_office_domestic_spaces}(a)) and unified navigation and person tracking in confined domestic spaces (Fig.~\ref{fig:real_world_office_domestic_spaces}(b)), as well as outdoor person following (Fig.~\ref{fig:real_world_deployment_outdoor}). Finally, Fig.~\ref{fig:real_world_outdoor_navigation_ablation} shows that real-world training enables the robot to reach the street lamp, whereas the
robot controlled by the Sim-only model remains in the dark area. Collectively, these results validate the practical value of UniTrackPLA: its unified panoramic policy enables a single model to perform omnidirectional person tracking and instruction-guided navigation across diverse real-world environments, while real-world supervision further strengthens sim-to-real generalization.

\section{Conclusion}
We presented UniTrackPLA, a unified panorama-language-action model for
instruction-guided navigation and dynamic person tracking. PAE preserves
temporal and azimuthal structure across panoramic views, while WAC models
action-conditioned future visual states to support consistency-guided
waypoint reuse and replanning. We also introduced OmniTrackNav-Bench, which
unifies simulated tracking, panoramic navigation, and real-world routes
under a shared observation and action space. Experiments demonstrate
consistent improvements over TrackVLA, while real-world deployment confirms
the benefits of panoramic perception and real-data training. Future work
will explore uncertainty-aware replanning, broader embodied tasks, and
efficient fully onboard deployment.

\bibliographystyle{IEEEtran}
\bibliography{bib}

\end{document}